\documentclass[10pt,twocolumn,letterpaper]{article}

\usepackage{wacv}              

\usepackage{graphicx}
\usepackage{amsmath}
\usepackage{amssymb}
\usepackage{booktabs}
\usepackage{bm}

\usepackage{pifont}
\newcommand{\xmark}{\ding{55}} 

\usepackage[pagebackref,breaklinks,colorlinks]{hyperref}

\usepackage[capitalize]{cleveref}
\crefname{section}{Sec.}{Secs.}
\Crefname{section}{Section}{Sections}
\Crefname{table}{Table}{Tables}
\crefname{table}{Tab.}{Tabs.}

\def\wacvPaperID{978} 
\def\confName{WACV}
\def\confYear{2025}

\begin{document}

\title{SplatFace: Gaussian Splat Face Reconstruction Leveraging an Optimizable Surface}

\author{Jiahao Luo$^{1}$ \qquad Jing Liu$^{2}$ \qquad James Davis$^{1}$ \\
$^{1}$University of California, Santa Cruz\quad $^{2}$ByteDance Inc.\\
\texttt{\{jluo53, davisje\}@ucsc.edu}
}

\maketitle

\begin{abstract}
  We present SplatFace, a novel Gaussian splatting framework designed for 3D human face reconstruction without reliance on accurate pre-determined geometry. Our method is designed to simultaneously deliver both high-quality novel view rendering and accurate 3D mesh reconstructions. We incorporate a generic 3D Morphable Model (3DMM) to provide a surface geometric structure, making it possible to reconstruct faces with a limited set of input images. We introduce a joint optimization strategy that refines both the Gaussians and the morphable surface through a synergistic non-rigid alignment process. A novel distance metric, splat-to-surface, is proposed to improve alignment by considering both the Gaussian position and covariance. The surface information is also utilized to incorporate a world-space densification process, resulting in superior reconstruction quality. Our experimental analysis demonstrates that the proposed method is competitive with both other Gaussian splatting techniques in novel view synthesis and other 3D reconstruction methods in producing 3D face meshes with high geometric precision.
\end{abstract}

\section{Introduction}
\label{sec:intro}




Human face models are used extensively in a variety of domains, such as 3D avatars \cite{zollhofer2011automatic}, biometric recognition\cite{blanz2003face}, photo editing \cite{yang2011expression}, and the movie industry\cite{borshukov2005universal}. Historically, the production of high-quality 3D facial models has required multi-view images and specialized equipment and setups \cite{debevec2000acquiring, borshukov2005universal}. Recent research has achieved impressive reconstruction results given just a few images, primarily through the integration of regression networks with a 3D Morphable Model (3DMM)\cite{bai2020deep, bai2021riggable, wu2019mvf, zielonka2022towards, feng2021learning}. However, mesh-based 3DMM face reconstruction focuses on geometric accuracy, often at the expense of photometric quality, ignoring complex scene elements and light interaction.

With the advent of Neural Radiance Fields (NeRF) \cite{mildenhall2021nerf}, facial capture has witnessed substantial progress. Geometry-aware NeRF-based methods have demonstrated their effectiveness in capturing complex facial movements \cite{Zielonka2022InstantVH} and rendering extrapolated views\cite{prinzler2023diner}. The introduction of 3D Gaussian Splatting (3DGS) \cite{kerbl3Dgaussians} shows considerable promise in tackling the dual challenges of achieving complex scene fidelity and speeding up both training and rendering. It employs an explicit, yet adaptive, representation\cite{zwicker2001surface} combined with differentiable rendering. Despite these advancements, NeRF and Gaussian splatting methods typically require either a substantial number of multi-view inputs such as video sequences and multi-view captures, or robust and accurate geometry guidance such as pre-determined and scene specific 3D depth information. These prerequisites can pose limitations in practical applications where such data is difficult to obtain. A method which requires only a few input images and no precise depth information would expand the range of circumstances in which face reconstruction is possible.

In this work, we introduce SplatFace, a novel Gaussian splatting framework designed for 3D human face reconstruction without the reliance on accurate pre-determined geometry. Our framework aims to simultaneously deliver high-quality novel view rendering and a precise mesh-based reconstruction. To achieve this, we incorporate a generic 3D Morphable Model (3DMM) for geometric guidance, allowing reconstruction with only a limited set of input images. We propose to jointly optimize Gaussians and the morphable model surface. This is achieved through a non-rigid alignment process that tightly integrates both components. We propose a novel splat-to-surface distance metric that enhances alignment by considering both Gaussian centers and their covariances. In addition, we augment the view-space adaptive densification technique of 3DGS, with world-space densification, capitalizing on the presence of the surface to enhance high-frequency skin detail capture. Our experimental evaluation demonstrates that the proposed method outperforms existing techniques in terms of both novel view synthesis image fidelity and 3D face mesh geometric accuracy.

%
%






\begin{figure*}[h!]
  \includegraphics[width=\textwidth]{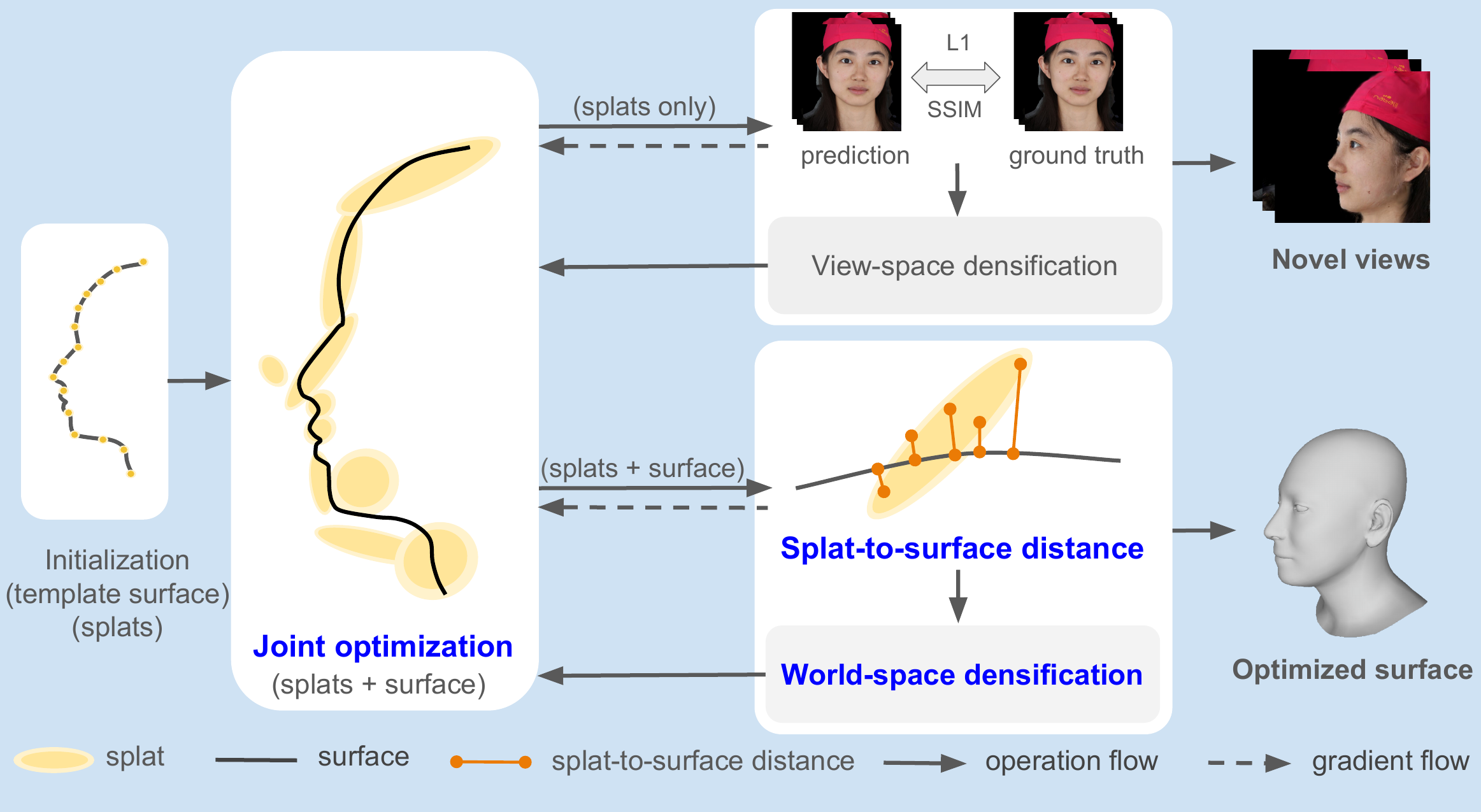}
  \caption{Diagram of overall process: SplatFace begins by initializing 3D Gaussians and a surface using the template mesh derived from a 3D Morphable Model (3DMM). The Gaussians and surface are then simultaneously refined through a \textit{joint optimization} process. Splats are constrained by image photoconsistency, and by a novel \textit{splat-to-surface distance} metric. This measure is introduced to accurately quantify the discrepancies between the Gaussian splat distribution and the surface, considering both the position and covariance of the Gaussians. Additionally, the presence of a surface allows the introduction of \textit{world-space densification}. As a result of this overall process, we obtain both enhanced novel view synthesis and a finely-tuned 3D mesh.}
  \label{method}
\end{figure*}

\section{Related Work}
\subsection{Few Shot 3D Face Reconstruction}

Realistic 3D human face reconstruction from a few 2D images has a long history of active research. Amazing results have been achieved with explicit representations such as meshes or depth maps \cite{deng2019accurate, xu2020deep, wu2019mvf, lei2023hierarchical, gecer2019ganfit, feng2021learning, luo2022much, luo2022accurate, kumar2023disjoint}. Typically 3D morphable face model parameters~\cite{blanz1999morphable} are regressed by training an encoder-decoder network~\cite{tuan2017regressing, zhu2016face}. There also exists work that directly regresses 3D geometries~\cite{guo2020towards, li2017learning, sanyal2019learning}. However, explicit representations face challenges in modeling complex scene elements such as hair, clothes or highly anisotropic surfaces. Neural radiance field (NeRF)~\cite{mildenhall2021nerf} based representations have also been used to represent a dynamic human head and body \cite{park2021nerfies, gao2022reconstructing, Zielonka2022InstantVH, prinzler2023diner}. 
Our work is directly inspired by this line of research, however we address few shot face reconstruction in the context of 3D Gaussian Splatting \cite{kerbl3Dgaussians}.

\subsection{Gaussian Splatting for Avatars}
The pursuit of real-time animatable volumetric avatar systems has been a challenging goal in virtual human research.  PointAvatar \cite{Zheng2023pointavatar} proposed a deformable point-based representation that achieves efficient training and re-rendering, even in new environments. 3D Gaussian Splatting \cite{kerbl3Dgaussians} has been quickly embraced by the community to build more efficient and animatable avatar systems. Recent methods \cite{dhamo2023headgas, zhao2024psavatar, xu2023gaussian, chen2023monogaussianavatar, xiang2024flashavatar} utilize monocular video inputs and generate new expressions by leveraging canonical space modeling and deformation prediction. GaussianAvatars \cite{qian2023gaussianavatars} takes multi-view video and binds the Gaussians onto mesh triangles, and optimizes the parameters of the Gaussian ellipsoids. Unfortunately, these methods require hundreds or thousands of multi-view frames of the target subject as input, in addition to pre-determined 3D geometry before splatting, in order to reconstruct a high fidelity avatar to animate. Besides, methods with monocular video as input tend to fail when there are changes in camera viewpoint, often resulting in numerous artifacts. In contrast, our work focuses on dramatically reducing the number of input frames to just a few images, enabling the accurate reconstruction of static 3D geometry and reliable rendering with changes in viewpoint.

\subsection{Few Shot Gaussian Splatting}
Despite the success of Gaussian Splatting in representing complex scenes, it remains a difficult problem to synthesize novel viewpoints when only a few input images are available. This condition is often called few-shot reconstruction and is difficult due to inherent ambiguities in learning 3D structure from a limited set of 2D images. Several methods based on Gaussian Splatting have been proposed to address the general scene few-shot reconstruction problem~\cite{xiong2023sparsegs, chung2023depth, zhu2023FSGS}. These methods leverage a pre-trained depth estimation model to extrapolate pseudo input views to assist the reconstruction process. Chung et al.~\cite{chung2023depth} add a depth map loss term together with a depth smoothness term to regularize the optimization process. SparseGS~\cite{xiong2023sparsegs} further incorporates a generative diffusion model to refine the pseudo input views, improving floater pruning and overall synthesis quality. FSGS~\cite{zhu2023FSGS} pairs depth regularization with a proximity-guided Gaussian unpooling process to enhance scene coverage and reconstruction.

These methods estimate a geometry prior in a pre-processing step which remains static during splat optimization, and use this prior primarily to constrain splat position. In contrast we use a geometry prior which is jointly optimized with the Gaussian splats, and is additionally used to guide both splat orientation and splat densification.


\section{Method}
The overall pipeline of our proposed method, SplatFace, is shown in Figure~\ref{method}. One key idea in our work is to insert an optimizable morphable surface which moves together with the Gaussian splats. The splats and surface are initialized using the same template surface across different identities. In contrast to 3DGS that focuses mainly on view-space operations and image similarity (the upper box showing operation and gradient flow), we propose a parallel process that focuses on world-space operations (the lower box showing operation and gradient flow). We propose \textit{joint optimization} of the Gaussians and surface using a \textit{splat-to-surface distance} metric, as well as \textit{world-space densification} to capture high-frequency details without generating extra floating splats. This process results in both a Gaussian splat model for novel view synthesis and an accurate 3D mesh representation of the surface. After a very brief introduction to Gaussian splatting, we provide a detailed presentation of each novel component in the subsequent sections.

\subsection{Background on Gaussian Splatting}

Gaussian splatting\cite{kerbl3Dgaussians} represents a 3D scene as a set
of 3D Gaussian primitives with adaptive covariances, and
renders an image using volume splatting. This model requires neither normal nor triangulation.  Each Gaussian splat is defined
by a full 3D covariance matrix $\Sigma$ centered at its position in the object space:

\begin{equation}
  G(\textbf{x}) = e^{-\frac{1}{2}(\textbf{x}-\boldsymbol{\mu})^{T}\Sigma^{-1}(\textbf{x}-\boldsymbol{\mu}))}.
  \label{eq:gaussian}
\end{equation}

Covariance matrices have physical meaning only when they are positive semi-definite. Gradient descent cannot be easily constrained to produce such valid matrices. Instead of directly optimizing the covariance matrix $\Sigma$ to obtain 3D Gaussians that represent the radiance field, we follow previous research\cite{zwicker2001surface} and approximate the covariance matrix with scaling and rotation:

\begin{equation}
  \Sigma = RSS^{T}R^{T}.
  \label{eq:covariance}
\end{equation}

To render images from new views, the color of a pixel is obtain by projecting and rasterizing Gaussian splats followed by alpha blending. Color of each splat is modeled with a view-dependent term using the following function:

\begin{equation}
  c = \sum_{j=1}^{n}c_{i}\alpha_{i}\prod_{j=1}^{i-1}(1-\alpha_{j}),
  \label{eq:covariance}
\end{equation}

where $\alpha_{i}$ and $c_{i}$ are the opacity and a spherical harmonic representation of color computed for each Gaussian splat.

For a more detailed description of Gaussian splatting we refer readers to prior research. Due to limited space, in this paper we focus on introducing and evaluating the novel contributions of our work.

\subsection{Joint Optimization of Gaussian Splats and Geometric Surface}



Our method results in a face reconstruction supporting novel view synthesis and producing a precise 3D geometric model. We propose a joint optimization between Gaussian splatting and a constrained surface model starting from a template shape.  This joint optimization is in contrast to existing work which relies on a pre-determined 3D geometry, such as monocular depth prediction\cite{zhu2023FSGS, xiong2023sparsegs}, multi-view stereo\cite{prinzler2023diner}, or head tracking \cite{dhamo2023headgas, zhao2024psavatar}.  



We adopt the popular blendshape 3D morphable model, FLAME\cite{li2017learning} as a generic representation of the face surface. This model is widely used in face reconstruction research so we do not discuss the details here. The model includes identity $\{\beta\}$, expression $\{\psi\}$ and pose $\{\theta\}$ parameters and is initialized to zero (identity to average face). The 3D morphable model representation has far fewer parameters than Gaussian splats, so is efficient to optimize and regularize. Gaussian positions $\{\bm{\mu}\}$ are initialized using the vertex of this generic surface template, Gaussian scale $\{\textbf{s}\}$ rotation $\{\textbf{r}\}$ are initialized the same way as 3DGS, and SH color $\{\textbf{c}\}$ are set to zero. During training, the Gaussians change and move according to photo-consistency with target images, but also follow a loss term for distance and orientation relative to the surface. The surface itself follows a loss which constrains it to lie near the position of Gaussian splats. Beyond this non-rigid alignment, we also add an overall scale $\{S\}$, rotation $\{R\}$ and translation $\{T\}$ on surface so that it can finetune the camera pose via a rigid transformation. To summarize, our method represents and jointly optimizes the 3D scene as a set of 3D Gaussians and a geometric surface, which can be parameterized as $\{\beta, \psi, \theta, \bm{\mu}, \textbf{s}, \textbf{r}, \textbf{c}, S, R, T\}$.




\begin{figure}[t]
  \includegraphics[width=\linewidth]{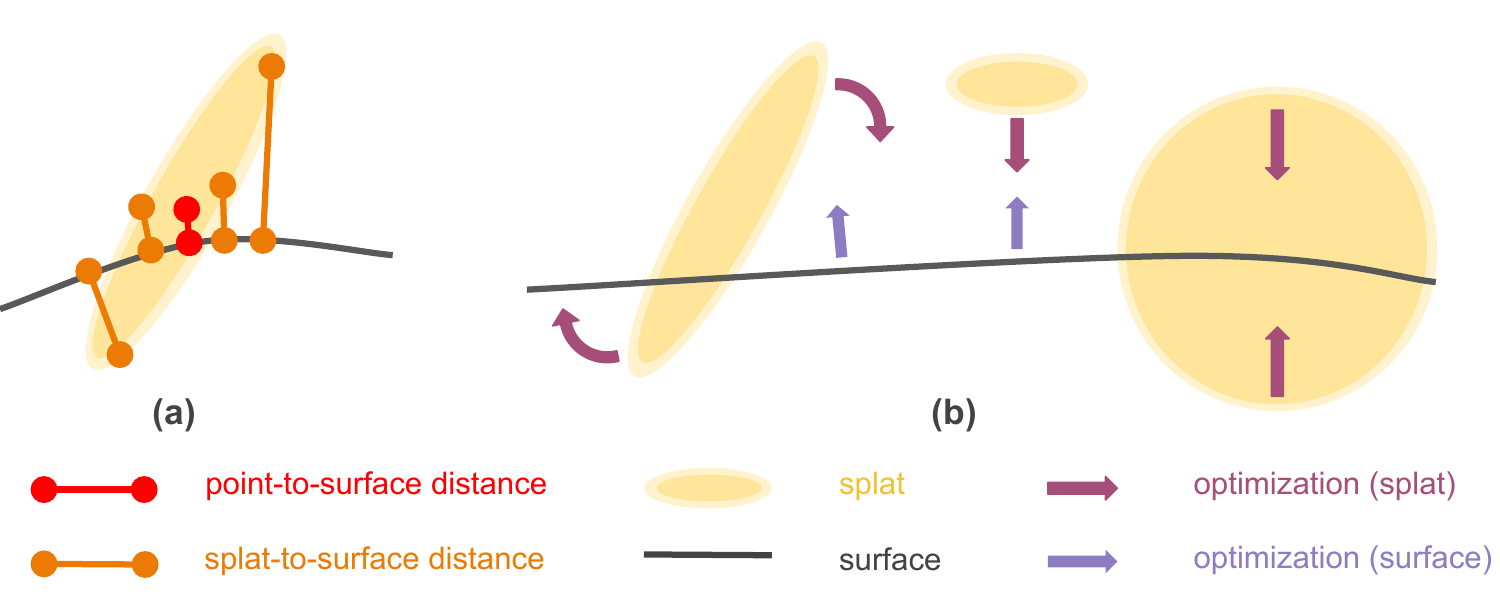}
  \caption{Joint optimization of Gaussians and geometric surface via splat-to-surface distance. (a) Illustration of splat-to-surface distance and (b) the modifications to splats and surface this distance is meant to encourage. Point-to-surface distance, shown in red, ignores covariance and considers each splat as a point. The point-to-surface distance will only calculate distance from splat center to the surface. In contrast, splat-to-surface distance accounts for the extended nature of the Gaussian distribution. Minimizing splat-to-surface distance will simultaneously optimize the orientation, position and scale of Gaussian Splats. }
  \label{s2f}
\end{figure}

\subsection{Splat-to-Surface Distance}


The joint optimization allows us to introduce a splat-to-surface distance measure. This distance can be used to enforce an additional loss during optimization intended to encourage the surface move toward Gaussians, while enforce Gaussians positioned near the surface rather than positioned away. From training viewpoints, Gaussians away from the surface may be effective at representing image variation, but from alternate viewpoints these Gaussians are perceived as floating artifacts. This problem is exacerbated when only a few viewpoints are available during training since insufficient image photoconsistency constraints exist to ensure  placement consistent with actual scene geometry.


To encourage Gaussians aligned with the surface, we introduce a term to minimize a distance function between the splats and the surface. Point-to-surface distance is the most commonly used distance functions to align point clouds to surfaces, and has been widely used in morphable model fitting and Iterative Closest Point (ICP). This measure is also known as point-to-face and point-to-plane. Given a point in the point cloud, point-to-surface distance is defined as the distance between the point and its closest triangle on the surface. Unfortunately, this widely used measure is designed for points with no spatial extent. Gaussian splats have a covariance matrix which defines a scale, and a more complex measure of distance is needed.

Figure~\ref{s2f}(a) shows a Gaussian splat positioned near to the surface. The splat position is indicated in red and the point-to-surface distance is the distance from its position (geometric center) to the surface, shown with a red bar. However, given a splat with scaling and rotation, a small distance between the splat center and the surface is not sufficient to guarantee a good alignment. In the example shown, some portions of the splat extend a considerable distance from the surface. To illustrate this point, several other positions sampled from the Gaussian distribution are shown with their distance to the surface shown with orange bars. When a splat is oriented as shown in this visualization it will lead to ``spiky'' artifacts when new views are rendered away from training viewpoints. 


To address this difficulty it is necessary to define a new distance measure which encourages splats not to extend too far from the surface. Figure~\ref{s2f}(b) shows the desired behavior. Splats should be encouraged to rotate, move towards the surface, and reduce size in the direction perpendicular to the surface. 

We propose a novel \textit{splat-to-surface} distance to correctly measure the distance between a splat distribution and the surface. Splat-to-surface distance can be calculated as the integral of the point-to-surface distance over the 3D Gaussian distribution as:

\begin{equation}
  \underset{\textbf{x}\sim G(\textbf{x})}{\int}G(\textbf{x}) \ |(\textbf{x}-\textbf{x}_{\textbf{i}}) \cdot \textbf{n}_{\textbf{i}}|,
  \label{eq:s2f_integral}
\end{equation}

where $\textbf{x}$ represent a position within the distribution.   $\textbf{x}_{\textbf{i}}$ is the corresponding point on the closest triangle, and $\textbf{n}_{\textbf{i}}$ is the unit normal at $\textbf{x}_{\textbf{i}}$. Making an assumption of local surface planarity, the dot product is the point-to-surface distance of each position within the Gaussian. $G(\textbf{x})$ is the 3D Gaussian distribution value defined in formula~\ref{eq:gaussian} which serves as the weights of each point-to-surface distance.

Since this distance measure must be estimated efficiently during optimization, we approximate the integral with random sampling. We sample from each splat's Gaussian distribution N times, and calculate the average of the sampled distances:

\begin{equation}
  \sum_{\textbf{x}_{\textbf{j}} \sim G(\textbf{x})}^{N} \ \frac{1}{N} \ |(\textbf{x}_{\textbf{j}}-\textbf{x}_{\textbf{i}}) \cdot \textbf{n}_{\textbf{i}}|,
  \label{eq:s2f_sum}
\end{equation}

where $\textbf{x}_{\textbf{j}}$ is one of the N samples.  In practice a small number of samples is sufficient because the sample position is randomized on each iteration, resulting in many samples over the course of optimization. We use N=2 for all examples, and provide an analysis in the supplemental material.

\subsection{World-space densification}



\begin{figure}[t]
  \includegraphics[width=\linewidth]{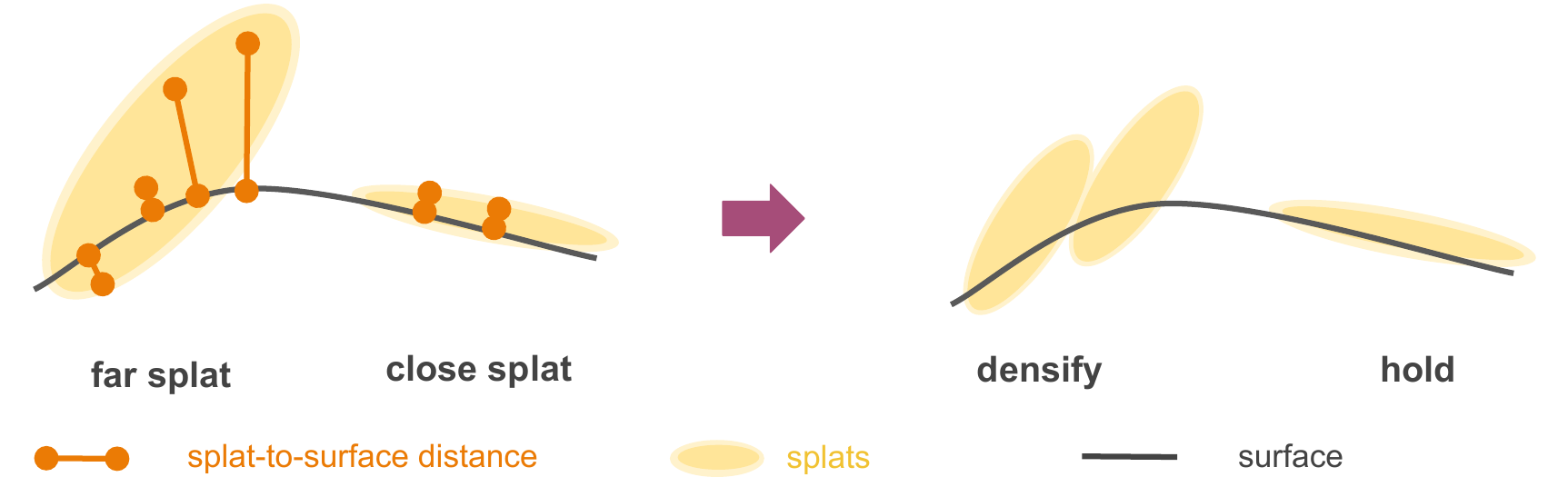}
  \caption{World-space densification.  We densify splats that are far from the surface using the proposed splat-to-surface distance. Gaussians with any sampled points far from the surface are considered far and either split or cloned. This helps generate multiple splats closer to the surface which are more expressive than a single floating or spiky splat which extends far from the surface.  }
  \label{densification}
\end{figure}

We propose world-space densification using the surface model to augment view-space densification with a new geometry based policy. Fig~\ref{densification} illustrate how world-space densification works. We densify splats with splat-to-surface distance further than an adaptive threshold $\tau_{s2s}$. When only a few input views are available, we observed that the Gaussian model tends to generate floaters and spikes with larger splat-to-surface distance to capture high-frequency details. It would be possible to increase the weight of the splat-to-surface loss to keep the Gaussian position and covariance very near the surface. However, it smooths out high-frequency skin details and results in over-regularization. We prefer more splats in those region for better detail capture, but these overfitted floaters or spikes with low RGB accumulated gradient in limited training view generally won't trigger view-space densification. A solution with a general lower view-space densification theshold causes densification in unnecessary regions. Another solution that simply increasing the weight of the RGB loss leads to insufficient geometric surface constraint, allowing more floaters and spikes. Thus, balancing the weights of the two losses and view-space densification threshold is neither sufficient nor effective.

We use the same densification techniques including clone and split from 3DGS. Splats which lie closer than the threshold distance are not densified by this rule, although they may be densified by view-space densification if they have a large accumulated RGB gradient. The densified floaters will be either dragged back to surface or pruned in later optimization. This approach maintains high-frequency skin detail capture while preventing artifacts in novel view synthesis. In addition, more splats in regions of complexity help the jointly optimized surface model to better capture the details of face shape.

\subsection{Combined loss function}
We optimize Gaussian splat parameters and morphable model parameters jointly with a total loss defined as: 


\begin{equation}
  loss = \lambda_{rgb} \ loss_{rgb} + \lambda_{s2s} \  loss_{s2s} + \lambda_{reg} loss_{reg},
  \label{eq:loss}
\end{equation}

where $loss_{rgb}$ is a combination of L1 and SSIM measuring RGB consistency on training views. This loss term is the same as vanilla 3DGS. $loss_{s2s}$ is our proposed splat-to-surface distance (Equation \ref{eq:s2f_sum}), measured as squared distance in meters, bringing splats and surface into alignment. We calculate $loss_{s2s}$ only in the face region since this is the only region in which the face model is valid. $loss_{reg}$ is a regularization term constraining surface shape to prevent overfitting. We empirically set $\lambda_{rgb}=1$, $\lambda_{s2s}=1000$ and $\lambda_{reg}=0.0001$. A sensitivity analysis of our new loss term, $\lambda_{s2s}$, is provided in the supplemental materials. In order to facilitate a fair comparison, we leave all other aspects of optimization identical to the baseline 3DGS implementation.

\section{Implementation, Results, and Evaluation}

\subsection{Implementation}

\subsubsection{Datasets} We evaluate our method on three datasets, FaceScape~\cite{yang2020facescape}, ILSH~\cite{zheng2023ilsh} and NeRSemble~\cite{kirschstein2023nersemble}. The FaceScape dataset contains high-resolution light-stage images captured by DSLR cameras on hundreds of identities. Each capture contains about 50 images with viewing angles spaced approximately 20 degree away from each other. The FaceScape dataset also provide 3D meshes, with reported reconstruction error less than 0.3mm. We randomly select 10 identities to evaluate both on 3D face mesh reconstruction and novel view synthesis.
The ILSH dataset include high-resolution light-stage capture from 24 cameras on 52 identities. We randomly select 10 identities to evaluate novel view synthesis.
The NeRSemble dataset include video recordings of 16 views covering the front and sides of the subject. We take the same frames from different views and randomly select 5 identities to evaluate novel view synthesis.

\subsubsection{Experimental Setting} We downsample all DSLR images (1086x724 in the FaceScape, 1024x750 in the ILSH dataset, 802x550 in the NeRSemble). We generate masks to remove background regions, and adopt dynamic background color. We adopt all hyper parameters of vanilla 3DGS and use our initialization for all baseline methods for a fair comparison. 
We train our proposed method with 10k iterations. Training takes about 10 minutes on a Nvidia RTX 3080.

\subsection{Comparison on surface mesh reconstruction}

Joint optimization of Gaussian splats \textit{and} 3D morphable model parameters results in a 3D mesh which estimates surface geometry. This 3D geometry is useful beyond view synthesis and many methods exist specifically to estimate 3D face shape from images. In order to evaluate our results, we compare with four state-of-the-art (SOTA) multi-view 3D face reconstruction methods: MVF-Net \cite{wu2019mvf}, DFNRMVS  \cite{bai2020deep}, INORig \cite{bai2021riggable} and HRN \cite{lei2023hierarchical} using the FaceScape dataset. HRN \cite{lei2023hierarchical} uses geometry disentanglement and introduces a hierarchical representation. MVF-Net \cite{wu2019mvf} and DFNRMVS  \cite{bai2020deep} train convolutional neural networks to explicitly enforce multi-view appearance consistency.  All of these methods focus on mesh generation, seeking to minimize 3D surface error rather than image based errors common in novel view synthesis. Since most of the comparison methods use 3 input views, we do the same for fair comparison.

For error analysis, the predicted meshes from each method are aligned to ground truth using the Iterative Closest Point (ICP) algorithm. For each point on the ground truth scan, we calculate the point-to-face distance in millimeters by finding the closest triangle in the predicted mesh. From this set of distances, we calculate summary statistics like mean-squared error (MSE), Median, and a robust approximation of maximum error which discards 10\% of high error points as outliers (M90).

\begin{figure}
  \centering
  \vspace{-10pt}
    \includegraphics[width=\linewidth]{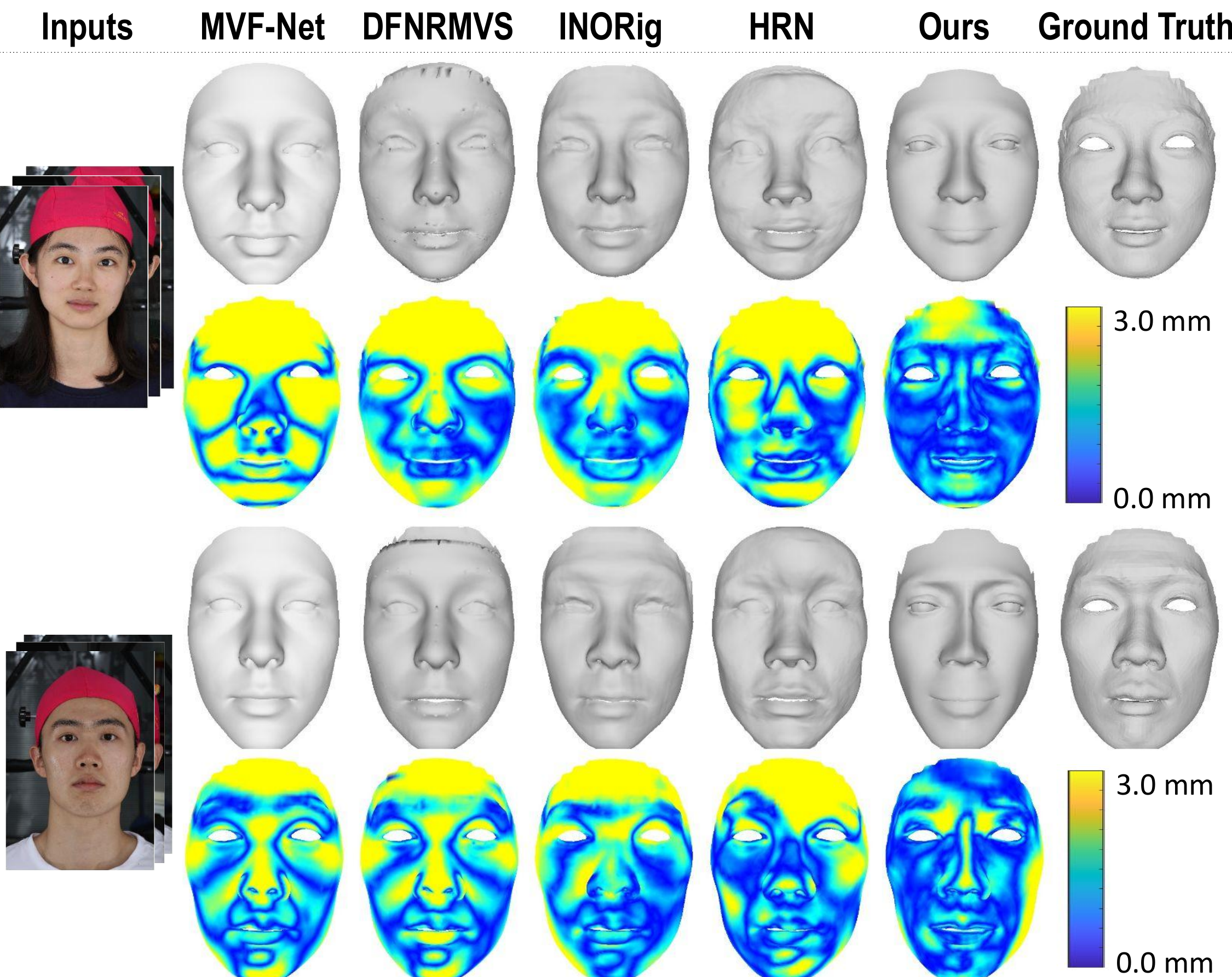}
    \caption{Qualitative comparison of face shape estimation with state of the art methods using the FaceScape dataset. The error maps visualize a range from blue (0 mm) to yellow (3 mm). Our method produces a smooth reconstruction with lower error.}
    \label{fig:mesh}
    \vspace{-10pt}
\end{figure}

Figure~\ref{fig:mesh} provide a qualitative comparison of the mesh and color map of mesh reconstruction error against the state-of-the-art dedicated 3D face estimation methods. Table~\ref{tab:reconstruction} presents a quantitative comparison. Our method outperforms the comparisons on all error metrics.

\begin{table}
\centering
\begin{tabular}{l|c|c|c}
\hline
Methods & MSE $\downarrow$ & Median $\downarrow$ & M90 $\downarrow$ \\
\hline
MVF-Net \cite{wu2019mvf}  & 1.75 & 1.45 & 3.60\\
DFNRMVS \cite{bai2020deep} & 1.73  & 1.40 & 3.56\\
INORig \cite{bai2021riggable} & 1.46  & 1.18 & 3.05\\
HRN \cite{lei2023hierarchical} & 1.28 & 0.96 & 2.54\\
Ours & \textbf{1.06} & \textbf{0.85} & \textbf{2.21}\\
\hline
\end{tabular}
\caption{Quantitative comparison of face shape estimation in FaceScape dataset. Geometric error in millimeters is provided using MSE, Median, and max error after rejecting 10\% outliers (M90). Our method outperforms existing few-view 3D face reconstruction methods. }
\label{tab:reconstruction}
\vspace{-10pt}
\end{table}

\begin{figure*}
  \includegraphics[width=\textwidth]{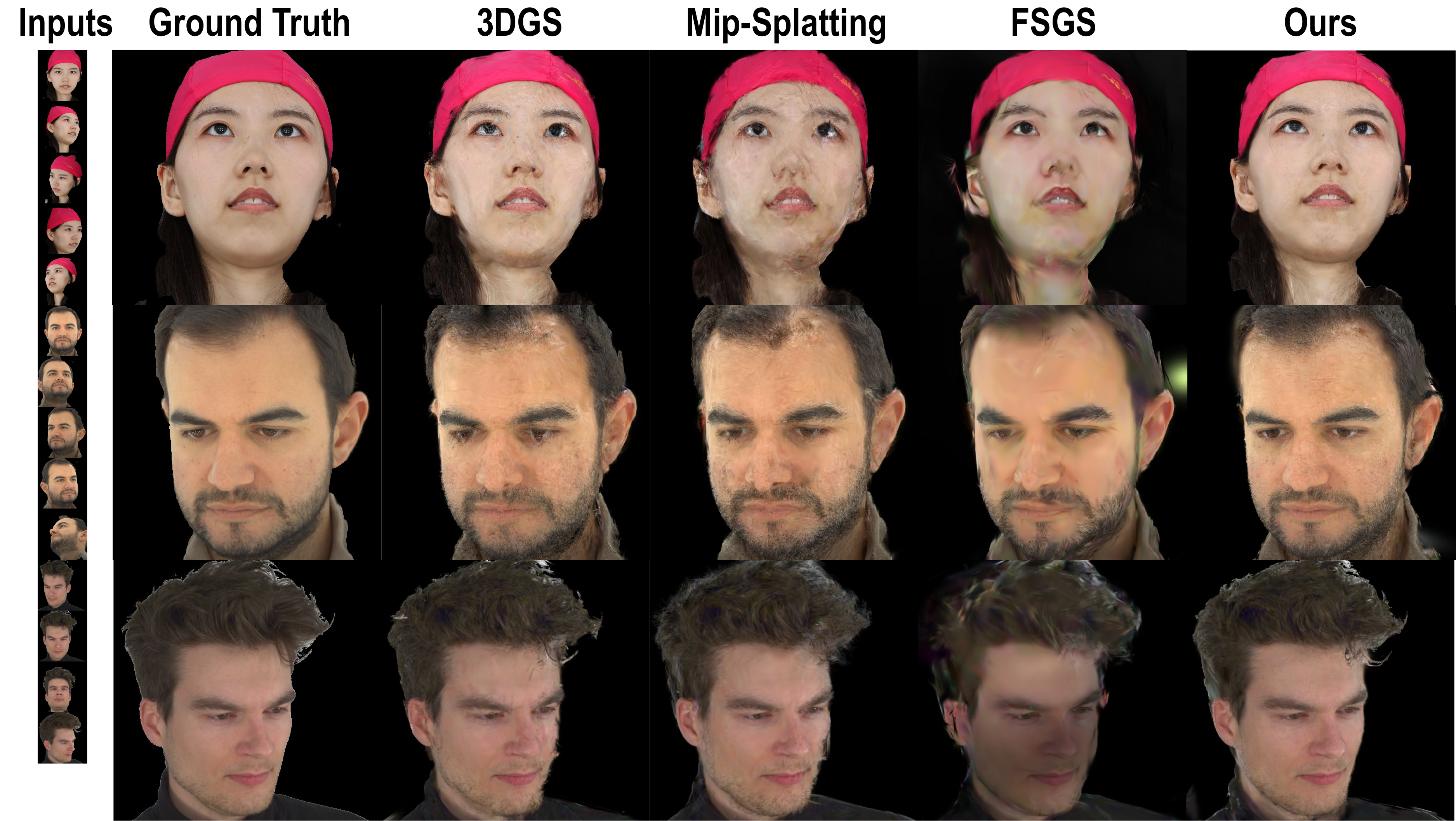}
  \caption{Qualitative comparison on novel view synthesis. We compare our results with other Gaussian splatting methods on the FaceScape\cite{yang2020facescape}, ILSH\cite{zheng2023ilsh} and NeRSemble\cite{kirschstein2023nersemble} datasets. Each method uses few-view input images, shown on the left. Our method produce results with fewer artifacts than the comparison methods.}
  \label{main}
\end{figure*}

\subsection{Comparison on novel view synthesis}


Figure~\ref{main} presents a qualitative visual comparison on novel view synthesis between our method and several baseline methods, including 3DGS~\cite{kerbl3Dgaussians}, Mip-splatting~\cite{yu2023mip} and FSGS~\cite{zhu2023FSGS}. We compare on the FaceScape, ILSH and NeRSemble datasets, using few images as input. Extrapolation of viewpoint away from training views is very challenging especially only a few views are available. 3DGS and Mip-splatting tend to produce noisy results and exhibit floating splats due to the lack of geometric constraints. FSGS often yields overly smoothed outcomes and results in mismatched colors and poor geometry, stemming from the use of a non-optimizable prior that lacks the precision necessary to properly constrain splat placement. While our method also contains artifacts, it yields the most visually appealing novel view synthesis.

Table~\ref{tab:main} presents a quantitative comparison of image error in terms of L1, SSIM, PSNR and LIPIPS on each dataset. Our method outperforms the comparison set of methods. Reported numbers are averaged across multiple views and across all test subjects in each dataset. 




\begin{table*}[htbp]
\centering
\resizebox{\textwidth}{!}{
\begin{tabular}{l|c|c|c|c|c|c|c|c|c|c|c|c}
\hline
 & \multicolumn {4}{|c|}{\textbf{FaceScape}} & \multicolumn {4}{|c}{\textbf{ILSH}} & \multicolumn {4}{|c}{\textbf{NeRSemble}} \\
\hline
Methods & L1 $\downarrow$ & SSIM $\uparrow$ & PSNR $\uparrow$ & LPIPS $\downarrow$ & L1 $\downarrow$ & SSIM $\uparrow$ & PSNR $\uparrow$ & LPIPS $\downarrow$ & L1 $\downarrow$ & SSIM $\uparrow$ & PSNR $\uparrow$ & LPIPS $\downarrow$ \\
\hline
3DGS \cite{wu2019mvf}  & 0.0273 & 0.8350 & 25.34 & 0.1404 & 0.0405 & 0.7277 & 22.04 &  0.2224 & 0.0231 & 0.8337 & 25.89 &  0.1354\\
Mip-Splatting \cite{bai2020deep} & 0.0436  & 0.8387 & 25.60 & 0.1379 & 0.0447 & 0.6826 & 22.18 &  0.2406 & 0.0228 & 0.8416 & 26.03 &  0.1276\\
FSGS \cite{bai2021riggable} & 0.0349 & 0.8307 & 24.27 & 0.2950 & 0.0492 & 0.7064 & 21.33 &  0.3535 &  0.0259 & 0.8391 & 25.20 &  0.1806\\
Ours & \textbf{0.0233} & \textbf{0.8556} & \textbf{26.58} & \textbf{0.1193} & \textbf{0.0391} & \textbf{0.7459} & \textbf{22.70} & \textbf{0.1871} & \textbf{0.0219} & \textbf{0.8621} & \textbf{26.72} &  \textbf{0.1181}\\
\hline
\end{tabular}
}
\caption{Quantitative results on the FaceScape, ILSH and NeRSemble datasets, using the L1, SSIM, PSNR and LPIPS image quality metrics. All metrics show that our method achieves better performance than the comparison Gaussian splatting methods.}
\label{tab:main}
\end{table*}

\begin{table}[h]
\centering
\resizebox{\columnwidth}{!}{
\begin{tabular}{l|c|c|c|c|c}
\hline
\begin{tabular}{@{}c@{}}Surface \\ initialization\end{tabular} & 
\begin{tabular}{@{}c@{}}Surface \\ optimization\end{tabular} & 
L1 $\downarrow$ & SSIM $\uparrow$ & PSNR $\uparrow$ & LPIPS $\downarrow$ \\
\hline
No surface & \xmark & 0.0273 & 0.8350 & 25.34 & 0.1404 \\
Template & \xmark & 0.0245 & 0.8495 & 25.79 & 0.1259\\
Template & \checkmark & \textbf{0.0233} & 0.8556 & 26.58 & \textbf{0.1193}\\
Ground truth & \checkmark & \textbf{0.0233} & \textbf{0.8716} & \textbf{26.68} & 0.1243\\
\hline
\end{tabular}
}
\caption{Ablation study showing the value of joint optimization and the acceptability of our surface initialization. A jointly optimized template surface results in higher quality images than the ablation conditions of no surface constraint or a non-optimized template surface. Initialization with a generic template surface reaches similiar image quality to initialization with the ground truth surface, indicating that optimization is able to adequately refine the surface representation.}
\label{tab:prior}
\vspace{-10pt}
\end{table}

\begin{table}[htbp]
\centering
\resizebox{\columnwidth}{!}{
\begin{tabular}{c|c|c|c|c|c|c}
\hline
\begin{tabular}{@{}c@{}}Point-to-surface \\ distance\end{tabular} & \
\begin{tabular}{@{}c@{}}Splat-to-surface \\ distance\end{tabular} & \ 
\begin{tabular}{@{}c@{}}World-space \\ densification\end{tabular} & L1 $\downarrow$ & SSIM $\uparrow$ & PSNR $\uparrow$ & LPIPS $\downarrow$ \\
\hline
\xmark & \xmark & \xmark & 0.0273 & 0.8350 & 25.34 & 0.1404\\
\checkmark & \xmark & \xmark & 0.0243 & 0.8492 & 26.14 & 0.1293\\
\checkmark & \xmark & \checkmark & 0.0237 & 0.8492 & 26.38 & 0.1249\\
\xmark & \checkmark & \xmark & 0.0239  & 0.8477 & 26.47 & 0.1243\\
\xmark & \checkmark & \checkmark & \textbf{0.0233} & \textbf{0.8556} & \textbf{26.58} & \textbf{0.1193}\\
\hline
\end{tabular}
}
\caption{Quantitative ablation study on splat-to-surface distance and world-space densification.  The proposed splat-to-surface distance metric results in higher image quality than than the ablation conditions of no surface constraint and a simple point-to-surface constraint. Incorporating world-space densification further refines performance. The highest image quality is obtained when using both of our enhancements.}
\label{tab:ab-methods}
\vspace{-10pt}
\end{table}

\subsection{Ablation study}



\subsubsection{Effectiveness of joint optimization and initialization} 
Our method jointly optimizes Gaussian splats and surface geometry parameters. Since splat locations are constrained by the surface, there is a concern that the surface might converge to an incorrect estimate of face shape and inadvertently drag the Gaussian splats to incorrect locations, degrading performance. To evaluate this concern we test our method both while holding the surface estimate constant and while allowing joint optimization. Table~\ref{tab:prior} provides a quantitative comparison of results using L1, SSIM, PSNR, and LPIPS as performance measures. As shown in the first three rows of the table, having a template surface estimate improves the rendered image quality compared to having no surface estimate, but jointly optimizing the splats and surface leads to even greater accuracy.


We initialize the surface prior to a template mesh representing the average face. This prior is only approximate initially, but is jointly optimized to find a high quality estimate of 3D face shape. Nevertheless, there may be residual shape error which affects Gaussian splat placement and thus image rendering. To assess the impact of using an approximation during surface initialization, we compare against using the ground truth face shape.  The final two rows of Table \ref{tab:prior} provide this comparison. Initialization with a template mesh produces renderings nearly as high quality as using a ground truth mesh for the surface. This indicates that our method is not very sensitive to initialization conditions, and thus using a simple template mesh is justified. 


\begin{figure*}
  \includegraphics[width=\textwidth]{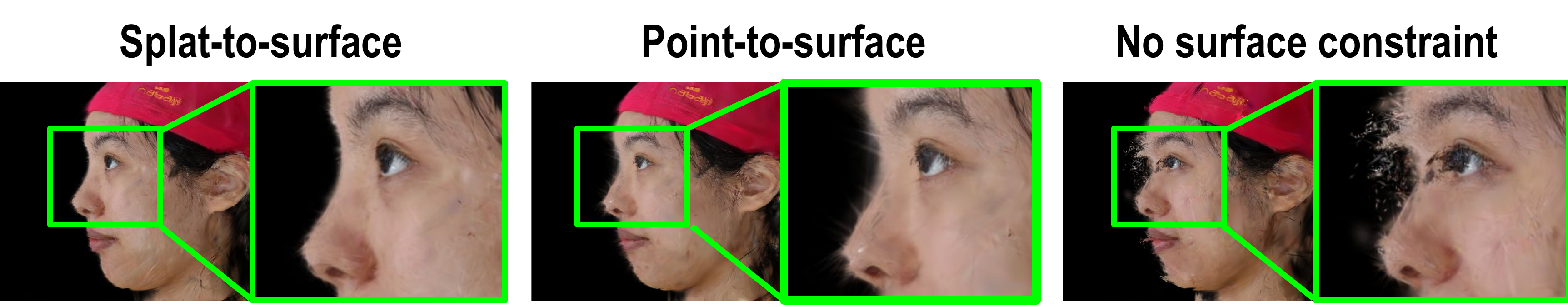}
  \caption{Visual ablation study on splat-to-face distance. When no surface constraint is present, many splats float away from the face surface. A simple point-to-surface constraint is effective in moving these splats to lie near to the surface, however the lack of an orientation constraint results in many ``spiky'' artifacts. Our splat-to-surface distance offers better alignment between splats and surface, resulting in fewer spiky artifacts in far test views.  }
  \label{ab_s2f}
  \vspace{-10pt}
\end{figure*}

\begin{figure}
  \includegraphics[width=\linewidth]{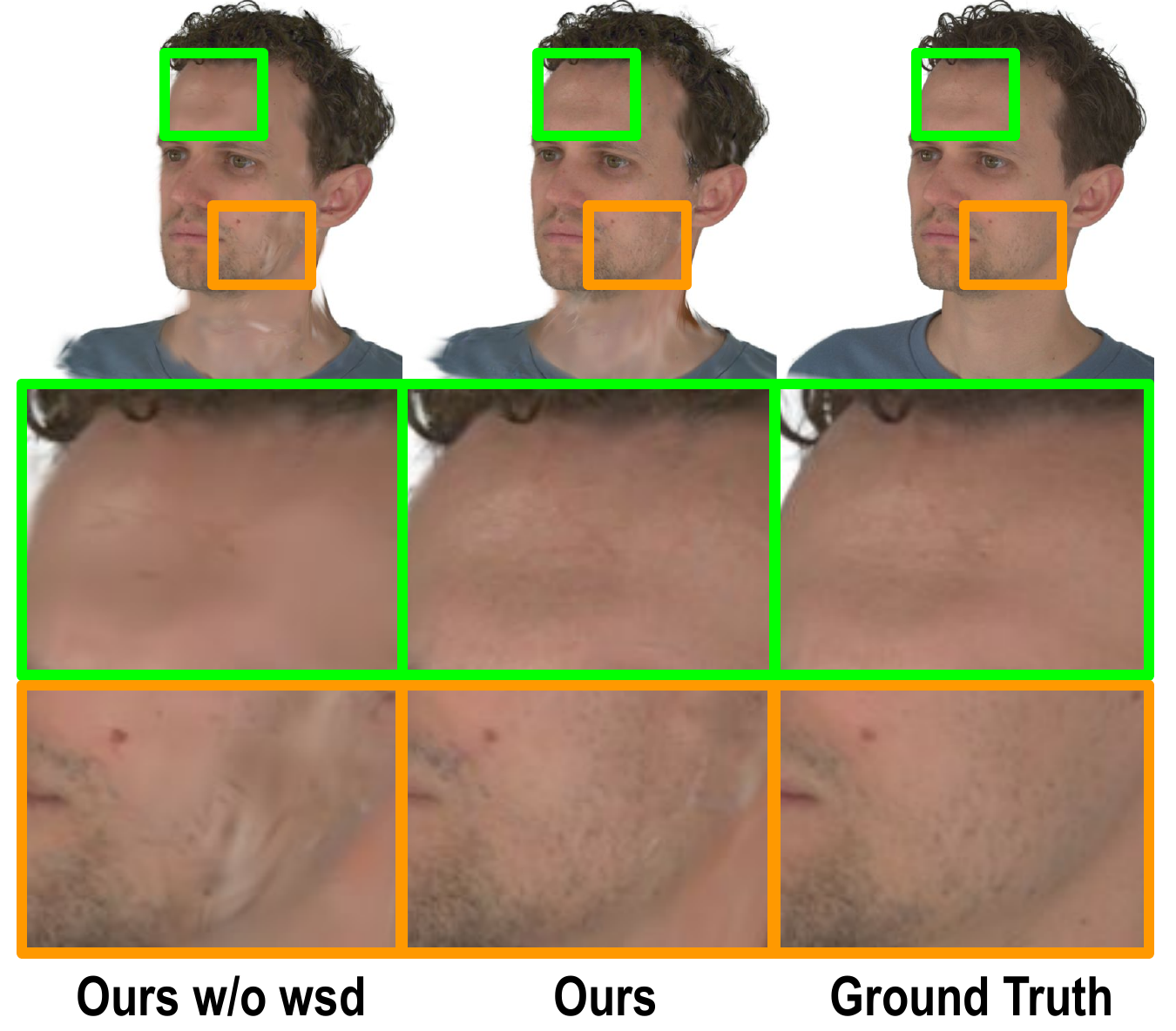}
  \caption{Visual ablation study on world-space densification (wsd). World-space densification offers better capture on high-frequency skin and facial hair details }
  \label{ab_wsd}
  \vspace{-10pt}
\end{figure}



\subsubsection{Effectiveness of splat-to-surface distance and world-space densification} The existence of a surface prior allows a splat-to-surface distance loss term as well as world space densification. We perform an ablation study to demonstrate the importance of these terms to our method.

Figure \ref{ab_s2f} illustrates the visual impact of splat-to-surface distance. The chosen test view is positioned at an angle greater than 40 degrees from any of the training views and emphasizes the artifacts that appear even at closer angles. Figure \ref{ab_wsd} illustrates the visual impact of world-space densification. Without it (w/o wsd) there is over-smoothing of features; adding it improves skin and facial hair details.

Table \ref{tab:ab-methods} presents novel view synthesis accuracy quantitatively, with and without our proposed components. Minimizing a distance function between splats and an optimizable surface enhances performance across all evaluation metrics. When contrasted with simple point-to-surface distance, our splat-to-surface approach exhibits superior improvement. We evaluated world space densification using both point-to-surface and splat-to-surface distance functions. Densification results in quantitative improvements in both cases.



\section{Limitations}

Our method aligns Gaussians to the surface, which can lead to over-regularization in regions where the surface model lacks the necessary expressiveness. Typical examples include complex geometries like teeth and accessories. In addition, our method only applies to scenes for which an optimizable surface models exist. We focus on face region in this work, and fall back to vanilla 3D Gaussian Splatting (3DGS) in non-face regions like hair and body. Finally, while our method outperforms existing Gaussian splatting methods, the rendered images are not yet completely free of artifacts. Achieving cleaner rendering, especially on far test views, remains a challenge.

\section{Conclusion}
In summary, we present SplatFace, an advancement in 3D human face reconstruction technology, which is particularly suited to the few-view input scenario. This  method avoids the need for precise predefined geometry, and instead jointly optimizes Gaussian splats and a morphable surface model. A splat-to-surface distance function and world-space densification serve to reduce artifacts caused by the limited number of input views. The method outputs both a Gaussian splatting model suitable for rendering novel viewpoints, and a 3D surface mesh suitable for traditional graphics pipelines including modeling and animation. Both of these outputs have been experimentally validated as higher accuracy than existing state of the art methods which focus on only one of the two goals.


%
%
{\small
\bibliographystyle{ieee_fullname}
\bibliography{paper}

\begin{thebibliography}{10}\itemsep=-1pt

\bibitem{bai2021riggable}
Ziqian Bai, Zhaopeng Cui, Xiaoming Liu, and Ping Tan.
\newblock Riggable {3D} face reconstruction via in-network optimization.
\newblock In {\em Proceedings of the IEEE/CVF Conference on Computer Vision and Pattern Recognition}, pages 6216--6225, 2021.

\bibitem{bai2020deep}
Ziqian Bai, Zhaopeng Cui, Jamal~Ahmed Rahim, Xiaoming Liu, and Ping Tan.
\newblock Deep facial non-rigid multi-view stereo.
\newblock In {\em Proceedings of the IEEE/CVF Conference on Computer Vision and Pattern Recognition}, pages 5850--5860, 2020.

\bibitem{blanz1999morphable}
Volker Blanz and Thomas Vetter.
\newblock A morphable model for the synthesis of {3D} faces.
\newblock In {\em Proceedings of the 22nd annual conference on Computer graphics and interactive techniques}, pages 351--358, 1995.

\bibitem{blanz2003face}
Volker Blanz and Thomas Vetter.
\newblock Face recognition based on fitting a {3D} morphable model.
\newblock {\em IEEE Transactions on Pattern Analysis and Machine Intelligence}, 25(9):1063--1074, 2003.

\bibitem{borshukov2005universal}
George Borshukov, Dan Piponi, Oystein Larsen, John~P Lewis, and Christina Tempelaar-Lietz.
\newblock Universal capture-image-based facial animation for" the matrix reloaded".
\newblock In {\em ACM Siggraph 2005 Courses}, pages 16--es. 2005.

\bibitem{chen2023monogaussianavatar}
Yufan Chen, Lizhen Wang, Qijing Li, Hongjiang Xiao, Shengping Zhang, Hongxun Yao, and Yebin Liu.
\newblock Monogaussianavatar: Monocular gaussian point-based head avatar.
\newblock {\em arXiv preprint arXiv:2312.04558}, 2023.

\bibitem{chung2023depth}
Jaeyoung Chung, Jeongtaek Oh, and Kyoung~Mu Lee.
\newblock Depth-regularized optimization for 3d gaussian splatting in few-shot images.
\newblock {\em arXiv preprint arXiv:2311.13398}, 2023.

\bibitem{debevec2000acquiring}
Paul Debevec, Tim Hawkins, Chris Tchou, Haarm-Pieter Duiker, Westley Sarokin, and Mark Sagar.
\newblock Acquiring the reflectance field of a human face.
\newblock In {\em Proceedings of the 27th Annual Conference on Computer Graphics and Interactive Techniques}, pages 145--156, 2000.

\bibitem{deng2019accurate}
Yu Deng, Jiaolong Yang, Sicheng Xu, Dong Chen, Yunde Jia, and Xin Tong.
\newblock Accurate {3D} face reconstruction with weakly-supervised learning: From single image to image set.
\newblock In {\em Proceedings of the IEEE/CVF Conference on Computer Vision and Pattern Recognition Workshops}, pages 0--0, 2019.

\bibitem{dhamo2023headgas}
Helisa Dhamo, Yinyu Nie, Arthur Moreau, Jifei Song, Richard Shaw, Yiren Zhou, and Eduardo P{\'e}rez-Pellitero.
\newblock Headgas: Real-time animatable head avatars via 3d gaussian splatting.
\newblock {\em arXiv preprint arXiv:2312.02902}, 2023.

\bibitem{feng2021learning}
Yao Feng, Haiwen Feng, Michael~J Black, and Timo Bolkart.
\newblock Learning an animatable detailed {3D} face model from in-the-wild images.
\newblock {\em ACM Transactions on Graphics (ToG)}, 40(4):1--13, 2021.

\bibitem{gao2022reconstructing}
Xuan Gao, Chenglai Zhong, Jun Xiang, Yang Hong, Yudong Guo, and Juyong Zhang.
\newblock Reconstructing personalized semantic facial nerf models from monocular video.
\newblock {\em ACM Transactions on Graphics (TOG)}, 41(6):1--12, 2022.

\bibitem{gecer2019ganfit}
Baris Gecer, Stylianos Ploumpis, Irene Kotsia, and Stefanos Zafeiriou.
\newblock Ganfit: Generative adversarial network fitting for high fidelity {3D} face reconstruction.
\newblock In {\em Proceedings of the IEEE/CVF Conference on Computer Vision and Pattern Recognition}, pages 1155--1164, 2019.

\bibitem{guo2020towards}
Jianzhu Guo, Xiangyu Zhu, Yang Yang, Fan Yang, Zhen Lei, and Stan~Z Li.
\newblock Towards fast, accurate and stable {3D} dense face alignment.
\newblock In {\em Computer Vision--ECCV 2020: 16th European Conference, Glasgow, UK, August 23--28, 2020, Proceedings, Part XIX}, pages 152--168. Springer, 2020.

\bibitem{kerbl3Dgaussians}
Bernhard Kerbl, Georgios Kopanas, Thomas Leimk{\"u}hler, and George Drettakis.
\newblock 3d gaussian splatting for real-time radiance field rendering.
\newblock {\em ACM Transactions on Graphics}, 42(4), July 2023.

\bibitem{kirschstein2023nersemble}
Tobias Kirschstein, Shenhan Qian, Simon Giebenhain, Tim Walter, and Matthias Nie{\ss}ner.
\newblock Nersemble: Multi-view radiance field reconstruction of human heads.
\newblock {\em ACM Transactions on Graphics (TOG)}, 42(4):1--14, 2023.

\bibitem{kumar2023disjoint}
Raja Kumar, Jiahao Luo, Alex Pang, and James Davis.
\newblock Disjoint pose and shape for 3d face reconstruction.
\newblock In {\em Proceedings of the IEEE/CVF International Conference on Computer Vision}, pages 3115--3125, 2023.

\bibitem{lei2023hierarchical}
Biwen Lei, Jianqiang Ren, Mengyang Feng, Miaomiao Cui, and Xuansong Xie.
\newblock A hierarchical representation network for accurate and detailed face reconstruction from in-the-wild images.
\newblock {\em Proceedings of the IEEE/CVF Conference on Computer Vision and Pattern Recognition}, 2023.

\bibitem{li2017learning}
Tianye Li, Timo Bolkart, Michael~J Black, Hao Li, and Javier Romero.
\newblock Learning a model of facial shape and expression from 4d scans.
\newblock {\em ACM Trans. Graph.}, 36(6):194--1, 2017.

\bibitem{luo2022much}
Jiahao Luo, Fahim~Hasan Khan, Issei Mori, Akila de Silva, Eric~Sandoval Ruezga, Minghao Liu, Alex Pang, and James Davis.
\newblock How much does input data type impact final face model accuracy?
\newblock In {\em Proceedings of the IEEE/CVF Conference on Computer Vision and Pattern Recognition}, pages 18985--18994, 2022.

\bibitem{luo2022accurate}
Jiahao Luo, Eric~Sandoval Ruezga, and James Davis.
\newblock How accurate is passive stereo for {3D} face reconstruction?
\newblock In {\em 2022 IEEE International Conference on Image Processing (ICIP)}, pages 2516--2520. IEEE, 2022.

\bibitem{mildenhall2021nerf}
Ben Mildenhall, Pratul~P Srinivasan, Matthew Tancik, Jonathan~T Barron, Ravi Ramamoorthi, and Ren Ng.
\newblock Nerf: Representing scenes as neural radiance fields for view synthesis.
\newblock {\em Communications of the ACM}, 65(1):99--106, 2021.

\bibitem{park2021nerfies}
Keunhong Park, Utkarsh Sinha, Jonathan~T. Barron, Sofien Bouaziz, Dan~B Goldman, Steven~M. Seitz, and Ricardo Martin-Brualla.
\newblock Nerfies: Deformable neural radiance fields.
\newblock {\em ICCV}, 2021.

\bibitem{prinzler2023diner}
Malte Prinzler, Otmar Hilliges, and Justus Thies.
\newblock Diner: Depth-aware image-based neural radiance fields.
\newblock In {\em Proceedings of the IEEE/CVF Conference on Computer Vision and Pattern Recognition}, pages 12449--12459, 2023.

\bibitem{qian2023gaussianavatars}
Shenhan Qian, Tobias Kirschstein, Liam Schoneveld, Davide Davoli, Simon Giebenhain, and Matthias Nie{\ss}ner.
\newblock Gaussianavatars: Photorealistic head avatars with rigged 3d gaussians.
\newblock {\em arXiv preprint arXiv:2312.02069}, 2023.

\bibitem{sanyal2019learning}
Soubhik Sanyal, Timo Bolkart, Haiwen Feng, and Michael~J Black.
\newblock Learning to regress {3D} face shape and expression from an image without {3D} supervision.
\newblock In {\em Proceedings of the IEEE/CVF Conference on Computer Vision and Pattern Recognition}, pages 7763--7772, 2019.

\bibitem{tuan2017regressing}
Anh Tuan~Tran, Tal Hassner, Iacopo Masi, and G{\'e}rard Medioni.
\newblock Regressing robust and discriminative {3D} morphable models with a very deep neural network.
\newblock In {\em Proceedings of the IEEE conference on Computer Vision and Pattern Recognition}, pages 5163--5172, 2017.

\bibitem{wu2019mvf}
Fanzi Wu, Linchao Bao, Yajing Chen, Yonggen Ling, Yibing Song, Songnan Li, King~Ngi Ngan, and Wei Liu.
\newblock Mvf-net: Multi-view {3D} face morphable model regression.
\newblock In {\em Proceedings of the IEEE/CVF Conference on Computer Vision and Pattern Recognition}, pages 959--968, 2019.

\bibitem{xiang2024flashavatar}
Jun Xiang, Xuan Gao, Yudong Guo, and Juyong Zhang.
\newblock Flashavatar: High-fidelity head avatar with efficient gaussian embedding.
\newblock In {\em Proceedings of the IEEE/CVF Conference on Computer Vision and Pattern Recognition}, pages 1802--1812, 2024.

\bibitem{xiong2023sparsegs}
Haolin Xiong, Sairisheek Muttukuru, Rishi Upadhyay, Pradyumna Chari, and Achuta Kadambi.
\newblock Sparsegs: Real-time 360° sparse view synthesis using gaussian splatting.
\newblock {\em Arxiv}, 2023.

\bibitem{xu2020deep}
Sicheng Xu, Jiaolong Yang, Dong Chen, Fang Wen, Yu Deng, Yunde Jia, and Tong Xin.
\newblock Deep 3d portrait from a single image.
\newblock In {\em Proceedings of the IEEE Conference on Computer Vision and Pattern Recognition (CVPR)}, 2020.

\bibitem{xu2023gaussian}
Yuelang Xu, Benwang Chen, Zhe Li, Hongwen Zhang, Lizhen Wang, Zerong Zheng, and Yebin Liu.
\newblock Gaussian head avatar: Ultra high-fidelity head avatar via dynamic gaussians.
\newblock {\em arXiv preprint arXiv:2312.03029}, 2023.

\bibitem{yang2011expression}
Fei Yang, Jue Wang, Eli Shechtman, Lubomir Bourdev, and Dimitri Metaxas.
\newblock Expression flow for {3D}-aware face component transfer.
\newblock In {\em ACM SIGGRAPH 2011 papers}, pages 1--10. 2011.

\bibitem{yang2020facescape}
Haotian Yang, Hao Zhu, Yanru Wang, Mingkai Huang, Qiu Shen, Ruigang Yang, and Xun Cao.
\newblock Facescape: A large-scale high quality {3D} face dataset and detailed riggable {3D} face prediction.
\newblock In {\em IEEE/CVF Conference on Computer Vision and Pattern Recognition (CVPR)}, June 2020.

\bibitem{yu2023mip}
Zehao Yu, Anpei Chen, Binbin Huang, Torsten Sattler, and Andreas Geiger.
\newblock Mip-splatting: Alias-free 3d gaussian splatting.
\newblock {\em arXiv preprint arXiv:2311.16493}, 2023.

\bibitem{zhao2024psavatar}
Zhongyuan Zhao, Zhenyu Bao, Qing Li, Guoping Qiu, and Kanglin Liu.
\newblock Psavatar: A point-based morphable shape model for real-time head avatar creation with 3d gaussian splatting.
\newblock {\em arXiv preprint arXiv:2401.12900}, 2024.

\bibitem{zheng2023ilsh}
Jiali Zheng, Youngkyoon Jang, Athanasios Papaioannou, Christos Kampouris, Rolandos~Alexandros Potamias, Foivos~Paraperas Papantoniou, Efstathios Galanakis, Ale{\v{s}} Leonardis, and Stefanos Zafeiriou.
\newblock Ilsh: The imperial light-stage head dataset for human head view synthesis.
\newblock In {\em Proceedings of the IEEE/CVF International Conference on Computer Vision}, pages 1112--1120, 2023.

\bibitem{Zheng2023pointavatar}
Yufeng Zheng, Wang Yifan, Gordon Wetzstein, Michael~J. Black, and Otmar Hilliges.
\newblock Pointavatar: Deformable point-based head avatars from videos.
\newblock In {\em Proceedings of the IEEE/CVF Conference on Computer Vision and Pattern Recognition (CVPR)}, 2023.

\bibitem{zhu2016face}
Xiangyu Zhu, Zhen Lei, Xiaoming Liu, Hailin Shi, and Stan~Z Li.
\newblock Face alignment across large poses: A {3D} solution.
\newblock In {\em Proceedings of the IEEE Conference on Computer Vision and Pattern Recognition}, pages 146--155, 2016.

\bibitem{zhu2023FSGS}
Zehao Zhu, Zhiwen Fan, Yifan Jiang, and Zhangyang Wang.
\newblock Fsgs: Real-time few-shot view synthesis using gaussian splatting, 2023.

\bibitem{Zielonka2022InstantVH}
Wojciech Zielonka, Timo Bolkart, and Justus Thies.
\newblock Instant volumetric head avatars.
\newblock {\em 2023 IEEE/CVF Conference on Computer Vision and Pattern Recognition (CVPR)}, pages 4574--4584, 2022.

\bibitem{zielonka2022towards}
Wojciech Zielonka, Timo Bolkart, and Justus Thies.
\newblock Towards metrical reconstruction of human faces.
\newblock In {\em European Conference on Computer Vision}, pages 250--269. Springer, 2022.

\bibitem{zollhofer2011automatic}
Michael Zollh{\"o}fer, Michael Martinek, G{\"u}nther Greiner, Marc Stamminger, and Jochen S{\"u}{\ss}muth.
\newblock Automatic reconstruction of personalized avatars from {3D} face scans.
\newblock {\em Computer Animation and Virtual Worlds}, 22(2-3):195--202, 2011.

\bibitem{zwicker2001surface}
Matthias Zwicker, Hanspeter Pfister, Jeroen Van~Baar, and Markus Gross.
\newblock Surface splatting.
\newblock In {\em Proceedings of the 28th annual conference on Computer graphics and interactive techniques}, pages 371--378, 2001.

\end{thebibliography}
}
\end{document}


\title{Supplementary Material}

\author{First Author\\
Institution1\\
Institution1 address\\
{\tt\small firstauthor@i1.org}
\and
Second Author\\
Institution2\\
First line of institution2 address\\
{\tt\small secondauthor@i2.org}
}
\maketitle

\section{Videos}

\begin{figure*}[t]
    \centering
    \includegraphics[width=\textwidth]{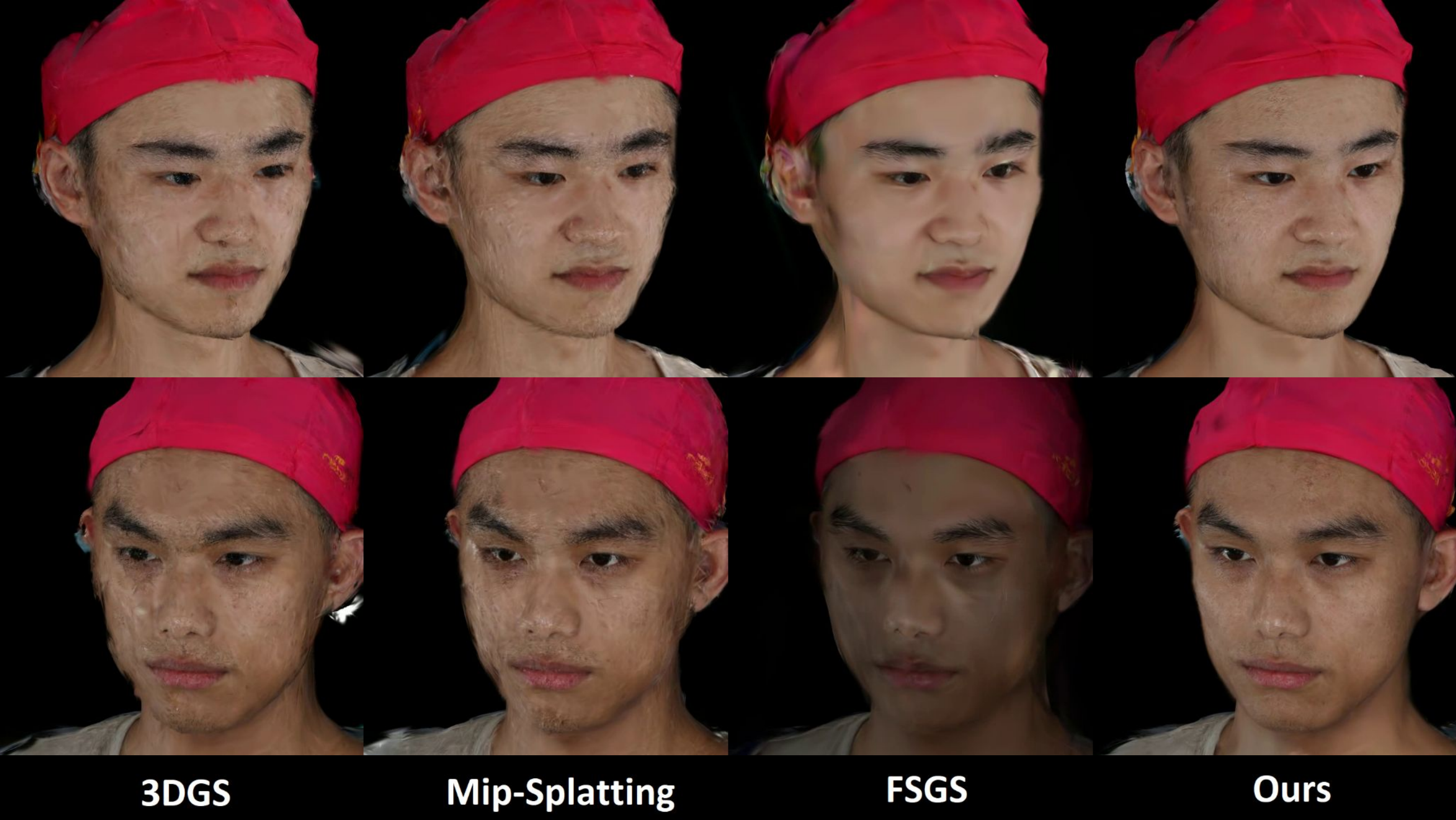}
    \caption{Sample video frame}
    \label{vid}
\end{figure*}

We includes 6 samples with various identities in the supplemental. It shows a comparison with other Gaussian splatting methods on the FaceScape dataset with 4-image inputs. We show an extrapolation from center to approximately 30 degree to the top left or top right. Figure \ref{vid} shows a qualitative comparison from two examples. 3DGS and Mip-Splatting generates floaters and spikes when the angle is large. FSGS generates overly smooth results and has lighting problems. Our method produces the most visually pleasing results with the fewest artifacts.

\begin{figure}[t]
  \centering
    \includegraphics[width=\linewidth]{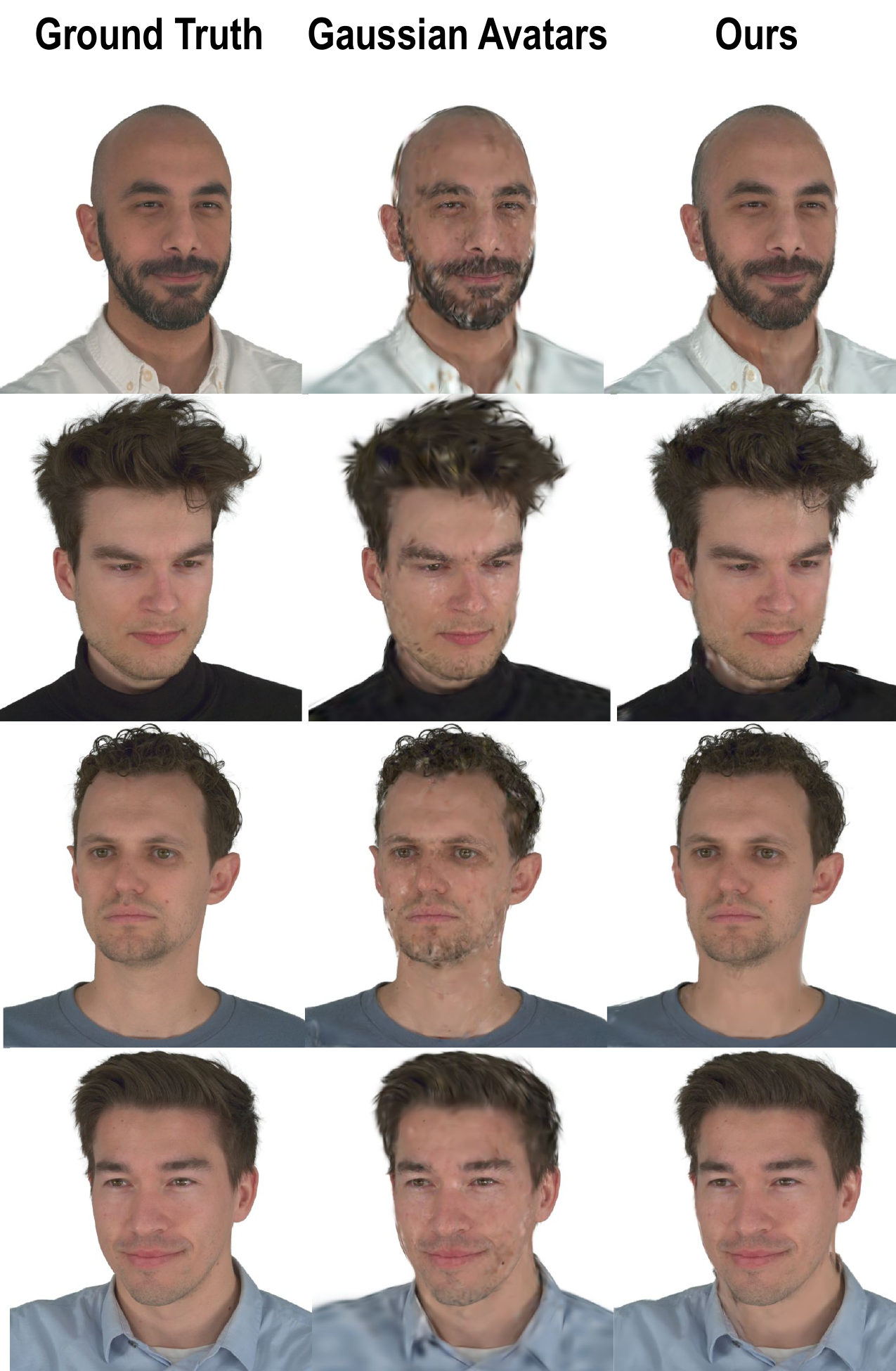}
    \caption{A qualitative comparison of novel view synthesis to Gaussian Avatars  with 4-view input.}
    \label{fig:GA}
\end{figure}

\section{Comparison with GaussianAvatar}
GaussianAvatar~\cite{qian2023gaussianavatars} is designed for animation (rather than static reconstruction) and with far more views (16 views rather than 3-5 views). It relies on pre-determined 3DMM parameters and only finetunes expression and pose. However, GaussianAvatar attaches Gaussians on the mesh triangles which is similar to our proposed non-rigid alignment. Therefore, we believe it's interesting to compare. GaussianAvatar regularizes position with a point-to-triangle distance that only uses the splat center, and regularizes covariance with a simple preference for smaller scale splats. This simpler loss works when many views are available, but breaks down with fewer views. Fig~\ref{fig:GA} provides a direct comparison to their provided implementation using NeRSemble data from their paper, restricted to 4 views. Our method produces results with fewer artifacts.

\section{Comparison with NeRF}

\begin{figure*}
  \includegraphics[width=\textwidth]{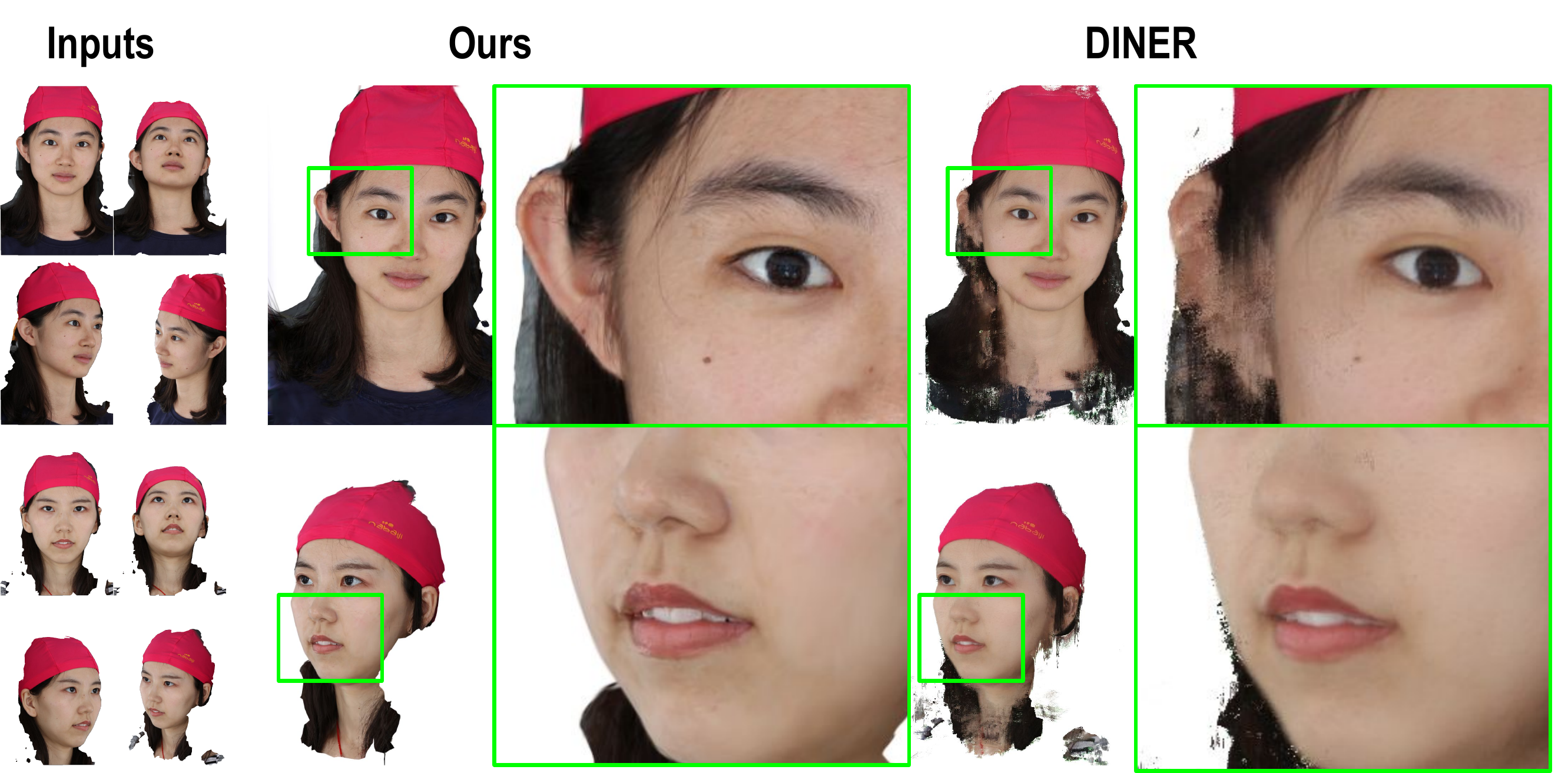}
  \caption{Qualitative comparison between our method and DINER on FaceScape dataset. The selected test view is close to the training views. Our method produce results with more high-frequency details and fewer artifacts.}
  \label{diner}
  \vspace{20pt}
\end{figure*}

Due to limited space in the main text, we showed only a comparison with other Gaussian Splatting methods. The NeRF family of techniques is closely related, so we provide here a comparison with the state-of-the-art 3D human face NeRF method, DINER\cite{prinzler2023diner}. This method was chosen because it's the most recent openly available method and specifically works well on 3D faces with only 4-image inputs. The authors reports higher performance than other few-view NeRF methods\cite{yu2021pixelnerf, mihajlovic2022keypointnerf}. Figure \ref{diner} shows a qualitative comparison between our method and DINER. Our method produce results with
more high-frequency details and fewer artifacts.

\begin{figure*}[t]
  \includegraphics[width=\textwidth]{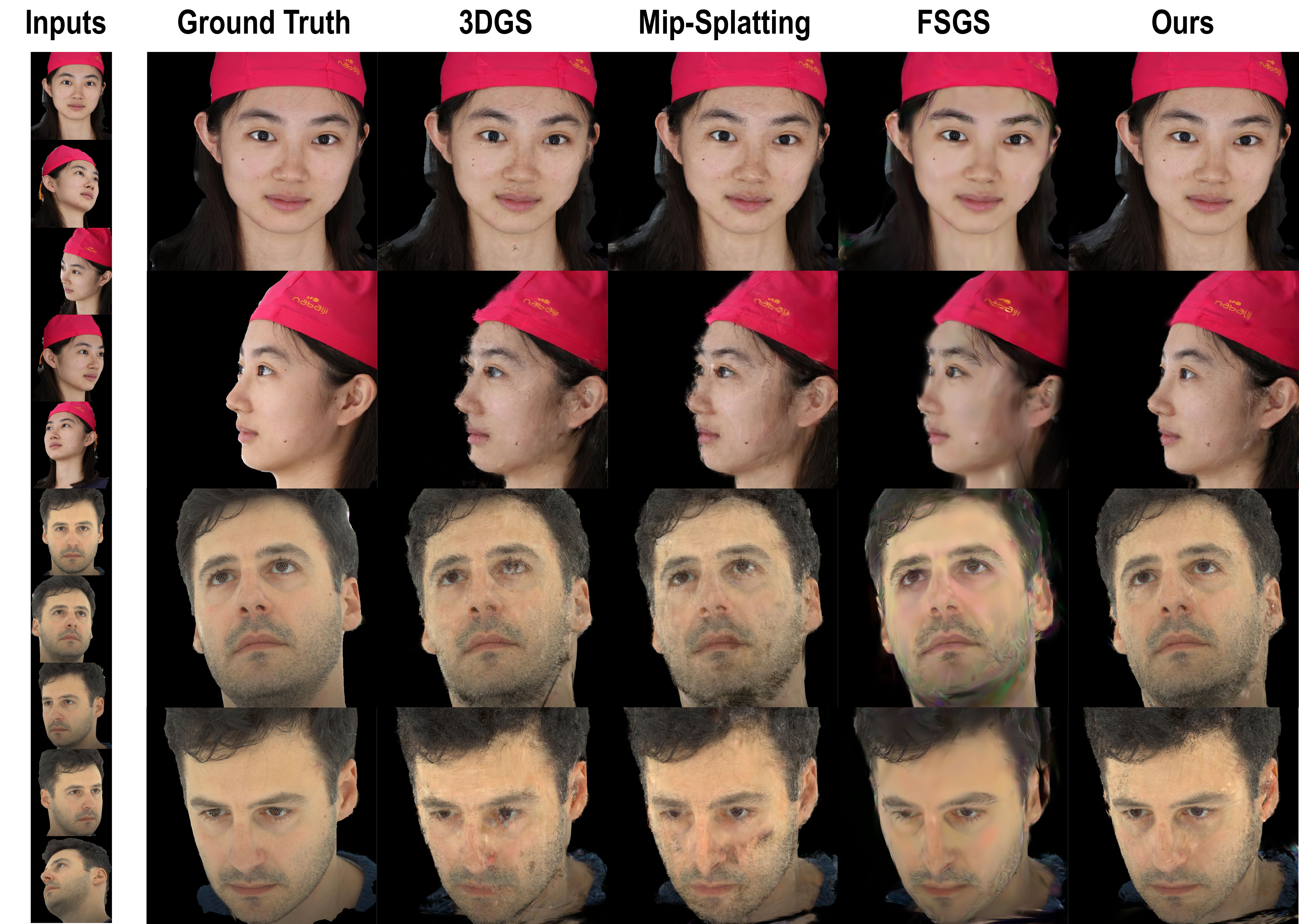}
  \caption{Qualitative comparison on novel view synthesis with different viewpoints. For each individual, a test view that is close to the training views is shown in the top row, and a test view further from the training images is shown in the bottom row. Our method produce results with fewer artifacts than the comparison methods.}
  \label{fig:viewpoint}
\end{figure*}

\section{Comparison with different viewpoints}
Figure~\ref{fig:viewpoint} shows an evaluation on both a near and far test viewpoint. In the first row of each individual, we show a test view near to one of the training views. 3DGS and Mip-splatting tend to produce noisy results, whereas FSGS often yields overly smoothed outcomes. In contrast, our method succeeds in capturing high-frequency details with minimal artifacts. In the second row for each test subject, we show an example which is far from the training views. Extrapolation of viewpoint far away from training views is very challenging and we do not expect perfect results from any method. 3DGS and Mip-Splatting produce noisy results and exhibit floating splats due to the lack of geometric constraints. These floating splats are most visible in profile views since they lie obviously away from the face. FSGS results in mismatched colors and poor geometry. While our method also contains artifacts, it yields the most visually appealing novel view synthesis. 

\section{Comparison with FlashAvatar}
\begin{figure}[t]
  \includegraphics[width=\linewidth]{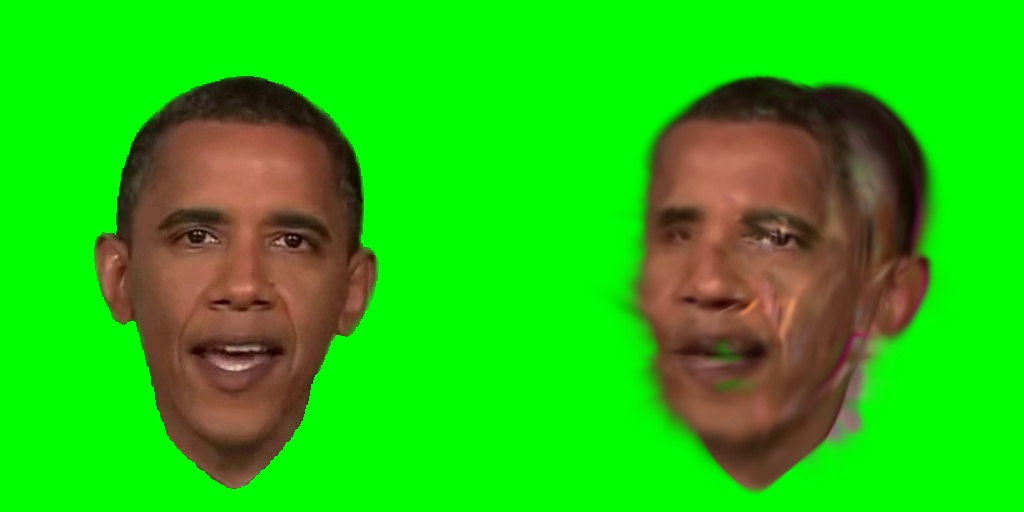}
  \caption{Monocular methods like FlashAvatar are not designed for novel view synthesis and thus perform poorly when rendering viewpoint is changed.}
  \label{fig:flashavatar}
\end{figure}

FlashAvatar~\cite{xiang2024flashavatar} is a method that uses a monocular video sequence during optimization. The results in that paper show animation from the same viewpoint as the training video. It is perhaps unfair to analyze the method on a scenerio it was not designed for. Nevertheless, in order to see if monocular input generalizes to novel viewpoints we include an example video while rotating the viewpoint. Figure~\ref{fig:flashavatar} shows one frame of this video. Notice that rendering quality is significantly degraded when the viewpoint is changed.

\section{Sensitivity Analysis of N}
\begin{figure}[t]
  \includegraphics[width=\linewidth]{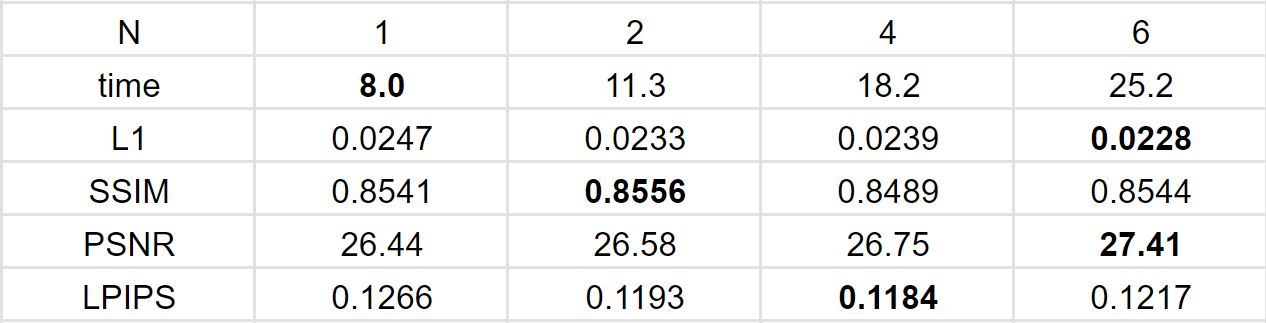}
  \caption{A sensitivity analysis of the effect of modifying the number of samples, N, used when sampling a Gaussian Splat reveals that the number of samples has only a mild effect on visual quality.}
  \label{fig:Nsensitivity}
\end{figure}

When sampling each Gaussian Splat to compute the SplatToSurface loss, the number of samples per iteration, N, has only a mild effect on visual quality. The samples are randomly chosen in each iteration, so each Gaussian is sampled thousands of times over the course of optimization. Figure~\ref{fig:Nsensitivity} shows a comparison on N=1,2,4,6, and lists optimization time in minutes, as well as four image quality metrics (L1, SSIM, PSNR, LPIPS). Notice that increased samples has little effect. All experiments in this paper were conducted with N=2. 

\section{Sensitivity Analysis of $\lambda_{s2s}$}
\begin{figure}[t]
  \includegraphics[width=\linewidth]{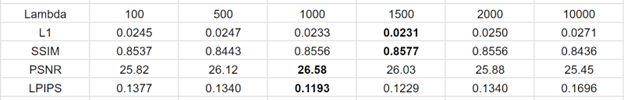}
  \caption{A sensitivity analysis of the effect $\lambda_{s2s}$, the parameter controlling the relative contribution of different loss terms. A relatively wide range of values produce adequate performance.}
  \label{fig:lambda}
\end{figure}

The relative weight of multiple loss terms in our method are controlled by the parameter $\lambda_{s2s}$. Figure~\ref{fig:lambda} shows multiple settings of this parameter and the effect on several measures of image quality. A relatively wide range of values are acceptable. All experiments in this paper use $\lambda$=1000.

{\small
\bibliographystyle{ieee_fullname}
\bibliography{egbib}
}